\documentclass[letterpaper]{article} 
\usepackage{aaai2026}  
\usepackage{times}  
\usepackage{helvet}  
\usepackage{courier}  
\usepackage[hyphens]{url}  
\usepackage{graphicx} 
\usepackage{natbib}  
\usepackage{caption} 
\usepackage{algorithm}
\usepackage{algorithmic}
\usepackage{amsmath}
\usepackage{booktabs}
\usepackage{multirow}
\usepackage{xcolor}
\usepackage[table]{xcolor}
\usepackage{tabularx}
\usepackage{amssymb}
\usepackage{newfloat}
\usepackage{listings}
\DeclareCaptionStyle{ruled}{labelfont=normalfont,labelsep=colon,strut=off} 
\floatstyle{ruled}
\newfloat{listing}{tb}{lst}{}
\floatname{listing}{Listing}
\title{\textsc{CallBench}: A Benchmark for Dual-Goal Coordination in Phone Call Assistants}

\author {
    Xuzhao Geng\textsuperscript{\rm 1},
    Haozhao Wang\textsuperscript{\rm 1},
    Xuelian Li\textsuperscript{\rm 2},
    Zhenyu Yang\textsuperscript{\rm 2},
    Haonan Lu\textsuperscript{\rm 2},\\
    Rui Zhang\textsuperscript{\rm 1},
    Ruixuan Li\textsuperscript{\rm 1}
}
\affiliations {
    \textsuperscript{\rm 1}School of Computer Science and Technology, Huazhong University of Science and Technology\\
    \textsuperscript{\rm 2}OPPO AI Center\\
    \{geng\_xz, hz\_wang, rxli\}@hust.edu.cn, \{lixuelian, yangzhenyu, luhaonan\}@oppo.com, rayteam@yeah.net
}

\usepackage{bibentry}
\nocopyright
\begin{document}

\maketitle

\begin{abstract}
Target-oriented dialogue systems have demonstrated strong capabilities in completing user goals through interactive conversations. However, existing studies are primarily designed for single, explicit goal completion, while phone call assistants face a proxy setting that requires coordinating the device owner's explicit preset goal with the caller's implicit and dynamic goal. We introduce \textsc{CallBench}, a Chinese benchmark for evaluating dual-goal coordination in phone call assistants. \textsc{CallBench} contains 50,000 complete multi-turn phone call dialogues across six scenarios: takeout, delivery, taxi, work, life, and harassment. It covers regular presets, emergent presets, and no-preset cases, and includes diverse relations between owner-side and caller-side goals, such as alignment, complementarity, irrelevance, and conflict. We further design a preset-aware turn-level evaluation protocol covering semantic understanding, context use, active guidance, response quality, preset compliance, dialogue rhythm, and safety. Experiments on representative dialogue methods show that existing approaches still struggle with this task, highlighting the need for phone call assistants that can make reliable turn-level decisions between two independent goals under proxy constraints.
\end{abstract}

\begin{figure}[t]
    \centering
    \resizebox{0.9\columnwidth}{!}{
    \includegraphics[width=\columnwidth]{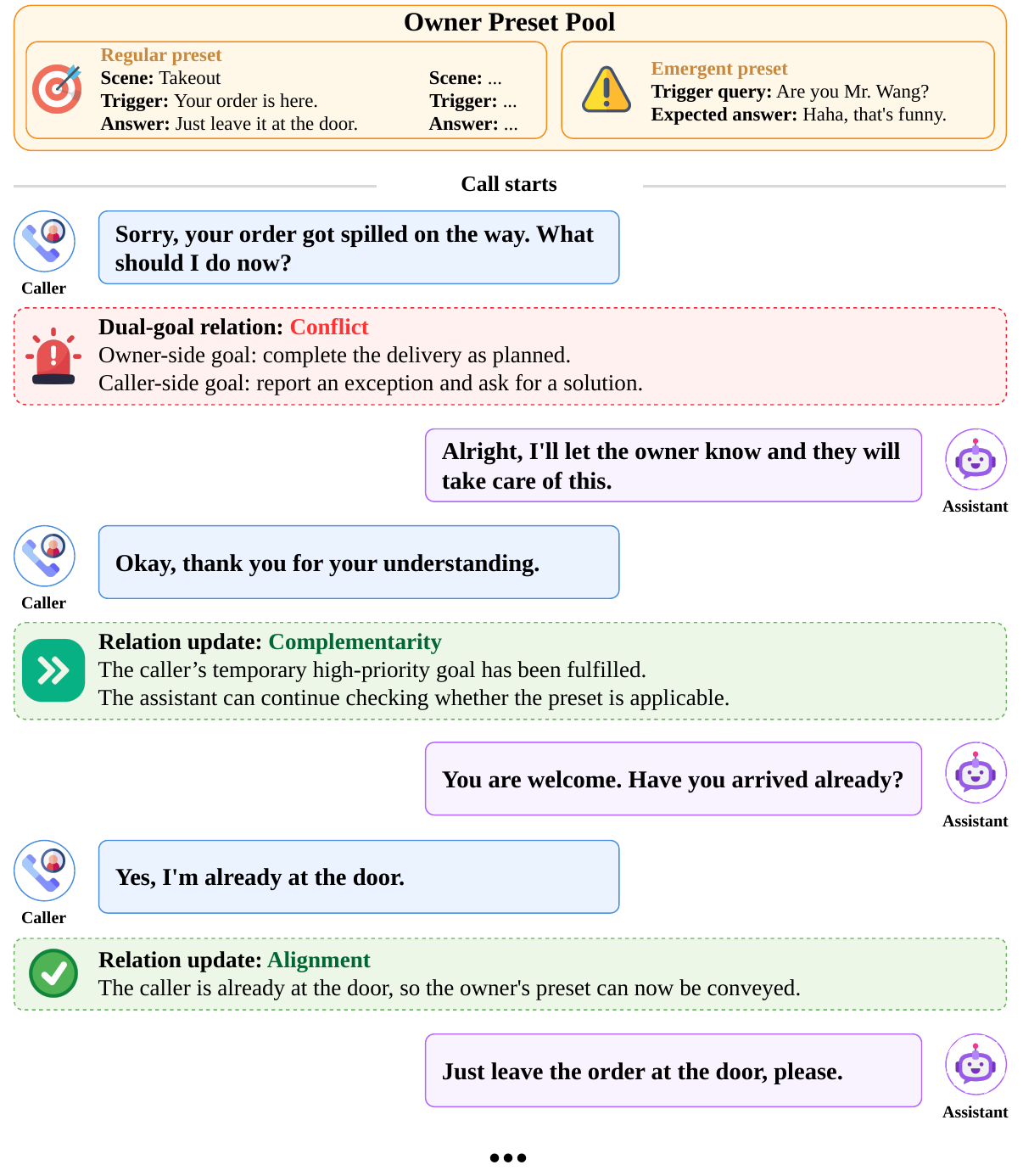}}
    \caption{A regular-preset example from \textsc{CallBench}.}
    \label{fig:example}
\end{figure}


\section{Introduction}

Target-oriented dialogue systems aim to guide multi-turn interactions toward predefined goals, such as recommendation, persuasion, or task completion \cite{lewis2017deal,wang2019persuasion,liu2021durecdial}. Existing studies have achieved strong progress in goal planning, response generation, and multi-turn decision-making \cite{deng2023survey}. However, most of them assume a relatively simple setting: the system directly interacts with the user whose goal should be completed. This single-goal assumption does not fully capture the proxy setting of phone call assistants.

A phone call assistant answers incoming calls on behalf of the device owner. Therefore, it must coordinate two different sides of goals: the owner's explicit preset and the caller's implicit, dynamic goal. These two goals may be aligned, complementary, irrelevant, or conflicting. For example, the owner may preset the instruction ``leave the takeout at the door'', while the caller reports that the food was spilled. Blindly advancing the preset in such cases may lead to an irrelevant or even harmful response. Thus, the key challenge is not merely completing a target, but deciding when the owner-side goal should be advanced, delayed, or suspended according to the evolving caller-side situation.

This proxy setting also introduces strict safety boundaries. The assistant must not reveal the owner's private information, fabricate unsupported facts, claim access to the physical world, or make decisions on behalf of the owner. In practice, the assistant can easily drift from reasonable assistance to unauthorized decision-making, such as agreeing to changes, making commitments, or refusing requests without the owner's confirmation. Therefore, safety is not an additional requirement after goal completion, but a fundamental condition for deploying phone call assistants.

Existing dialogue benchmarks are insufficient for evaluating these abilities. Traditional task-oriented dialogue datasets mainly focus on intent recognition, dialogue state tracking, API calls, and task success under user-centered goals \cite{budzianowski2018multiwoz,rastogi2020towards}. Target-oriented dialogue benchmarks emphasize goal advancement, but usually assume a single stable target \cite{lewis2017deal,wang2019persuasion,liu2021durecdial,deng2023survey}. Recent LLM-based dialogue and agent benchmarks evaluate general response quality, tool use, or policy compliance \cite{zheng2023judging,yao2024tau,li2023api}, yet they rarely consider a proxy assistant that must mediate between two independent persons under strict safety constraints. As a result, we still lack a benchmark for evaluating dual-goal coordination in phone call assistants.

To address this gap, we introduce \textsc{CallBench}, a Chinese benchmark for evaluating \textit{Dual-Goal Coordination} in phone call assistants. Although real call assistants operate through voice interfaces, most practical systems follow an ASR--text processing--TTS pipeline. We therefore focus on the text-based dialogue processing stage, which allows us to isolate decision-making and response-generation abilities from speech recognition and synthesis errors. \textsc{CallBench} contains 50,000 complete multi-turn phone call dialogues across six scenarios: takeout, delivery, taxi, work, life, and harassment. The benchmark covers regular presets, emergent presets, and no-preset cases, and includes diverse relations between owner-side and caller-side goals. Figure~\ref{fig:example} shows a regular-preset takeout example in \textsc{CallBench}, where the assistant must first handle a caller-side exception before returning to the owner-side preset.

We further design a preset-aware turn-level evaluation protocol for phone call assistants. Instead of relying only on task success or surface-level response quality, our evaluation measures semantic understanding, context use, active guidance, response quality, preset compliance, dialogue rhythm, and safety. This allows \textsc{CallBench} to reveal whether an assistant can make appropriate local decisions under dual-goal conditions, such as conveying a preset at the right time, handling caller-side exceptions, avoiding premature or repeated preset delivery, and staying within proxy-specific safety boundaries.

Our main contributions are as follows:
\begin{itemize}
    \item We formulate the call assistant task as a turn-level decision process, dynamically coordinating the owner's explicit targets and the caller's implicit intentions under strict safety constraints.
    \item We construct \textsc{CallBench}, a Chinese multi-scenario benchmark containing 50,000 complete multi-turn phone call dialogues across six scenarios, three preset cases, and diverse dual-goal relations.
    \item We deploy a fine-grained evaluation framework to assess model capabilities across multiple aspects.
    \item We evaluate representative task- and target-oriented dialogue frameworks, quantifying critical bottlenecks in dual-goal trade-offs, exception handling, and boundary control to guide future research.
\end{itemize}

\section{Related Work}

\subsection{Task- and Target-oriented Dialogue Benchmarks}

Task-oriented dialogue benchmarks have been widely used to evaluate dialogue state tracking, intent recognition, and task completion. Representative datasets such as MultiWOZ \cite{budzianowski2018multiwoz}, SGD \cite{rastogi2020towards}, and CrossWOZ \cite{zhu2020crosswoz} focus on multi-domain slot filling and API-oriented task execution. Other datasets further introduce more realistic interaction settings, including transactional dialogues in Taskmaster \cite{byrne2019taskmaster} and AirDialogue \cite{wei2018airdialogue}, implicit preference elicitation in Decision-Oriented Dialogue \cite{lin2024decision}, and negotiation in CaSiNo \cite{chawla2021casino}. These benchmarks have advanced the evaluation of multi-turn dialogue systems under explicit or partially implicit user goals.

However, most of these benchmarks assume that the system directly serves the current interactor. The system's role is usually to understand and complete the user's own task, rather than act as a proxy for another person. This assumption does not match phone call assistants, where the assistant must coordinate the device owner's explicit preset with the caller's dynamic implicit goal. Existing benchmarks provide limited coverage of cases where the two goals are irrelevant, complementary, or conflicting, and they rarely treat proxy-specific constraints such as privacy protection and unauthorized decision-making as central evaluation criteria.

\subsection{Policy-Constrained Agentic Dialogue Evaluation}

Recent LLM-agent benchmarks extend dialogue evaluation beyond text generation to tool use, policy compliance, and multi-turn decision-making \cite{schick2023toolformer,liu2024agentbench}. For example, \(\tau\)-bench \cite{yao2024tau} and \(\tau^2\)-bench \cite{barres2025tau} evaluate agents in customer-service environments with policy documents and controlled interactions. ABCD \cite{chen2021action} aligns customer requests with institutional policies and action sequences, while ToolTalk \cite{farn2023tooltalk} studies safety issues in tool-use dialogues, especially under irreversible actions. Related work such as PAChat \cite{fu2025pachat} further explores multi-speaker and personalized communication settings.

These benchmarks cover important aspects of policy following, action correctness, and communication structure, but they still differ from the proxy setting of phone call assistants. Their constraints are usually defined by institutions or tools, while the interaction remains largely between a user and a service system. In contrast, a phone call assistant mediates between two independent natural persons: the device owner and the caller. It must decide when to advance, delay, or suspend the owner's preset while respecting the caller's current need and strict safety boundaries. \textsc{CallBench} targets this gap by evaluating dual-goal coordination under phone-call-specific constraints, including goal conflict handling, preset timing, privacy protection, and decision-boundary compliance.

\section{Problem Definition}\label{sec:problem_definition}

Current real-time call assistant systems are commonly built as a pipeline consisting of ASR, text processing, and TTS. This paper focuses on the text processing stage. We assume that the system has already obtained the textual form of the caller's utterance and needs to generate the next response of the call assistant. Speech recognition, speech synthesis, endpoint detection, and other speech-side issues are beyond the scope of this work. Under this setting, we define the task of \textit{Dual-Goal Coordination} for real-time call assistants.

Given the dialogue history at turn $t$:
\[
H_t=\{(u_1,a_1),(u_2,a_2),\ldots,(u_{t-1},a_{t-1})\},
\]
where $u_i$ denotes the caller's utterance at turn $i$, and $a_i$ denotes the assistant's response at turn $i$. The current input includes the caller's latest utterance $u_t$, a scene label $s\in\mathcal{S}$, and the owner's preset pool $\mathcal{P}^o$, where:
\[
\mathcal{S}=\{\text{takeout},\text{delivery},\text{taxi},\text{work},\text{life},\text{harassment}\}.
\]

The owner's preset pool $\mathcal{P}^o$ contains goal-related information predefined by the device owner. To align with practical phone call assistant systems under deployment, we model $\mathcal{P}^o$ as a structured preset interface rather than an arbitrary set of instructions. Each preset can be represented as:
\[
p_i^o=(q_i^o,r_i^o),
\]
where $q_i^o$ denotes the trigger expression or trigger condition, and $r_i^o$ denotes the corresponding owner-specified response. Each preset slot may also be left empty.

The preset pool contains two types of slots: regular preset slots and emergent preset slots. Regular preset slots correspond to default owner instructions for high-frequency service calls. In our setting, regular preset slots are only defined for \textit{takeout}, \textit{delivery}, and \textit{taxi}. Each of these scenes has at most one regular preset slot. For a regular preset, the trigger expression $q_i^o$ is predefined by the system, and the owner only specifies the response $r_i^o$.

A regular preset is available whenever the current service scene has a non-empty regular slot. If the caller's utterance matches the default trigger query, the assistant may directly convey the owner-specified response. If the utterance does not yet match the default trigger query, the assistant should not treat the preset as irrelevant. Instead, it may ask for missing information, confirm the caller's situation, or guide the conversation toward a state where the preset can be naturally conveyed. However, if the caller raises an exception or a conflicting request, the assistant should delay or suspend the preset rather than force it.

Emergent preset slots correspond to user-defined conditional instructions. For an emergent preset, both the trigger expression $q_i^o$ and the response $r_i^o$ are specified by the owner. Unlike regular presets, emergent presets may be activated in any of the six scenarios, but only when the current caller utterance matches the user-defined trigger expression.

At turn $t$, the assistant retrieves the owner-side explicit goal available in the current context:
\[
g_t^o=\textsc{RetrievePreset}(H_t,u_t,s,\mathcal{P}^o).
\]
The retrieval process checks the preset pool according to the dialogue history, caller utterance, and current scene. For \textit{takeout}, \textit{delivery}, and \textit{taxi}, the assistant first checks the regular preset slot associated with the current scene. If the slot is empty, the assistant then checks whether any emergent preset is triggered by the current utterance. For \textit{work}, \textit{life}, and \textit{harassment}, the assistant only checks whether the current caller utterance matches any user-defined emergent trigger condition. If no preset is available, then $g_t^o=\varnothing$.

This setting gives rise to three preset cases. A \textit{\textbf{regular preset}} case occurs when the current service scene has a non-empty regular preset slot. An \textit{\textbf{emergent preset}} case occurs when a user-defined trigger condition is matched by the caller's utterance. A \textit{\textbf{no-preset}} case occurs when no owner-side preset is available or activated in the current turn. Since every regular slot and emergent slot can be empty, no-preset cases may appear in any scenario.

In contrast to the owner-side explicit goal, the caller's goal at the current turn is denoted as $g_t^c$. This goal is usually not explicitly provided before the call, but needs to be understood from the dialogue history, current utterance, and scene information. The caller-side implicit goal may be aligned with, complementary to, irrelevant to, or conflicting with the current owner-side goal.

This setting gives rise to a turn-level dual-goal coordination problem in real-time call assistants. At each turn, the system needs to determine the relationship between the current owner-side explicit goal $g_t^o$ and the caller-side implicit goal $g_t^c$, and generate the assistant response:
\[
a_t=f(H_t,u_t,s,g_t^o,g_t^c;\mathcal{C}),
\]
where $\mathcal{C}=\{C_{\text{priv}},C_{\text{dec}},C_{\text{fact}},C_{\text{phys}}\}$ denotes the set of safety boundary constraints that the call assistant must follow. Specifically, $C_{\text{priv}}$ denotes the privacy protection constraint, requiring the system not to disclose the owner's personal information, location, contact information, or other sensitive content; $C_{\text{dec}}$ denotes the decision-boundary constraint, requiring the system not to make commitments, agreements, refusals, or choices on behalf of the owner; $C_{\text{fact}}$ denotes the factuality constraint, requiring the system not to fabricate information that is not provided in the dialogue context or the owner's presets; and $C_{\text{phys}}$ denotes the non-embodiment constraint, requiring the system not to present itself as an embodied agent with direct sensory access or physical agency in the real world, such as claiming that it can see an object, check a location, or physically handle an item. The response $a_t$ should advance or complete the owner-side goal at an appropriate time, while also properly addressing the caller's current request. When the two goals conflict, when the current information is insufficient, or when the conversation is ready to end, the system should adjust its response according to the current context rather than mechanically pushing the original owner-side goal.

Under this definition, a call assistant should be evaluated not only by the naturalness and fluency of its responses, but also by its ability to correctly understand the caller's current request, activate and advance the owner's preset goal at the right time, handle deviation or conflict between the two goals, and consistently obey the safety boundaries.

\section{Benchmark Construction}

Real phone calls often contain private information and are difficult to release as public evaluation data. To enable controlled and reproducible evaluation, we construct \textsc{CallBench}, a Chinese benchmark for \textit{Dual-Goal Coordination} in phone call assistants. In total, \textsc{CallBench} contains 50,000 multi-turn Chinese phone call dialogues across six incoming-call scenarios, covering different preset cases, caller-side goals, owner-side goals, and dual-goal relations.

\subsection{Scenario and Preset Configuration}

\textsc{CallBench} covers the six incoming-call scenarios defined above, spanning short service calls and more open-ended interpersonal calls. In \textit{takeout}, \textit{delivery}, and \textit{taxi}, calls are usually short and revolve around a concrete service task, such as delivering an order, placing a package, or coordinating a ride. In \textit{work}, \textit{life}, and \textit{harassment}, caller intents are more diverse, requiring the assistant to retain useful information, respond cautiously, or terminate the call safely.

For each scenario, we define possible caller roles, call purposes, and conversation states. In \textit{takeout}, \textit{delivery}, and \textit{taxi}, callers may report normal arrival, address ambiguity, item damage, failed delivery, or ask for confirmation. In \textit{work} and \textit{life}, callers may ask for the owner, leave a message, provide new information, or request a decision. In \textit{harassment}, callers may engage in telemarketing, repeatedly ask irrelevant questions, try to obtain private information, or pressure the assistant into making unauthorized commitments.

Following the preset cases defined in the problem definition, \textsc{CallBench} includes regular presets, emergent presets, and no-preset instances. Regular presets are included only in \textit{takeout}, \textit{delivery}, and \textit{taxi}, where stable owner instructions commonly exist. Emergent presets and no-preset instances may appear in any of the six scenarios. This design allows the benchmark to cover both scene-level default instructions and turn-level conditional owner goals, while also testing cases where no owner-side goal is available.

\subsection{Dual-Goal Relation Design}

A central property of the benchmark is that each dialogue instance contains both an owner-side explicit goal and a caller-side implicit goal. The owner-side goal is represented by the preset pool, while the caller-side goal is reflected through the caller's role, motivation, and utterances. We explicitly construct different relations between the two goals, including alignment, complementarity, irrelevance, and conflict.

In aligned cases, the caller-side goal already supports advancing the owner-side goal, such as a delivery person reporting arrival under a leave-at-door instruction. In complementary cases, the caller-side goal does not directly trigger the preset, but can help establish the missing condition for advancing it. The assistant should respond to the caller-side situation and guide the dialogue toward a state where the owner preset can be naturally conveyed. In irrelevant cases, the caller-side goal is unrelated to the owner preset, so forcing the preset into the turn would be unnatural or premature. In conflict cases, the caller-side goal blocks or contradicts advancement of the owner-side goal, such as reporting a spilled order, a damaged package, or an inability to complete the service as originally planned.

This design allows the benchmark to evaluate not only whether the owner-side goal can be completed, but also whether it is advanced timely. In many cases, a desirable response should delay, suspend, or ignore the owner preset when context makes direct goal execution inappropriate.

\subsection{Dialogue Construction}

\begin{table*}[t]
\centering
\begin{tabular}{lcccccccc}
\toprule
\multirow{2}{*}{\textbf{Scene}} 
& \multicolumn{2}{c}{\textbf{Overall}} 
& \multicolumn{2}{c}{\textbf{Regular Preset}} 
& \multicolumn{2}{c}{\textbf{Emergent Preset}} 
& \multicolumn{2}{c}{\textbf{No Preset}} \\
\cmidrule(lr){2-3}
\cmidrule(lr){4-5}
\cmidrule(lr){6-7}
\cmidrule(lr){8-9}
& \textbf{Dialogues} & \textbf{Avg. Turns} 
& \textbf{Dialogues} & \textbf{Avg. Turns} 
& \textbf{Dialogues} & \textbf{Avg. Turns} 
& \textbf{Dialogues} & \textbf{Avg. Turns} \\
\midrule
Takeout     & 8,333 & 4.812 & 5,000 & 5.156 & 1,667 & 4.318 & 1,666 & 4.273 \\
Delivery    & 8,333 & 4.437 & 5,000 & 4.388 & 1,667 & 4.833 & 1,666 & 4.188 \\
Taxi        & 8,334 & 4.985 & 5,000 & 5.333 & 1,667 & 4.406 & 1,667 & 4.521 \\
Work        & 8,334 & 4.892 & --    & --    & 4,167 & 5.031 & 4,167 & 4.753 \\
Life        & 8,333 & 5.567 & --    & --    & 4,170 & 5.518 & 4,163 & 5.616 \\
Harassment  & 8,333 & 7.241 & --    & --    & 4,167 & 7.067 & 4,166 & 7.416 \\
\midrule
Total       & 50,000 & 5.322 & 15,000 & 4.959 & 17,505 & 5.485 & 17,495 & 5.471 \\
\bottomrule
\end{tabular}
\caption{Statistics of \textsc{CallBench}. We report the number of dialogues and the average turns per dialogue.}
\label{tab:data_statistics}
\end{table*}

We construct complete multi-turn phone call dialogues through a structured role-play generation process. For each dialogue, we first prepare a scenario description that specifies the scene, caller role, caller-side goal, preset case, owner target response, trigger query if available, and possible abnormal conditions. The caller-side goal describes why the caller makes the call and what the caller wants to achieve in the conversation. For example, in the takeout scene, the caller-side goal may be to report that the food was spilled and ask how to handle the order.

We use Doubao-Seed-1.6 \cite{bytedance2025seed16} to implement two role-playing agents, one acting as the caller and the other acting as the phone call assistant. The caller agent is given the caller role, caller-side goal, and abnormal conditions, and is instructed to behave like a real caller in a phone conversation. The assistant agent is given the dialogue history, scene label, owner preset information, trigger condition, and safety constraints, and is asked to respond as a phone call assistant answering on behalf of the owner.

Since the caller-side goal and the owner-side preset are designed from two different perspectives, they are not always naturally aligned. Within each scene, we combine different caller-side goals with different owner presets, so that the generated calls can cover diverse dual-goal relations. 

Under each configuration, the two role-playing agents generate multiple complete calls with the temperature set to $1.0$. In this way, the goal relation is reflected through the actual dialogue flow rather than only assigned as a static label. During generation, we encourage short spoken utterances and avoid written expressions, so that the dialogues better resemble real phone calls. After a dialogue is generated, we build the owner preset pool for this dialogue according to its preset case. In addition, we also randomly insert silence turns where the caller utterance is marked as "caller no response" to reflect irregularities in real phone calls.

\textsc{CallBench} stores each generated sample as a complete multi-turn dialogue. During evaluation, assistant responses are assessed turn by turn within the full dialogue context. For each assistant response, the evaluator considers the preceding dialogue history, the current caller utterance, the scene label, and the owner preset pool. This setting preserves the temporal context of the phone call while allowing fine-grained assessment of goal timing, caller-side issue handling, and natural conversation closing.

\subsection{Quality Control}

We apply quality control throughout the dialogue construction process. The automatic checking modules are implemented with Doubao-Seed-2.0-lite \cite{bytedance2026seed2}. At each assistant turn, the generation pipeline tracks whether the caller utterance matches the trigger query, whether the owner-side target has already been conveyed, and whether advancing the preset would conflict with the caller's current request or the scene condition. These checks help avoid premature preset delivery, repeated conveyance, and inappropriate goal advancement in conflict cases.

Candidate assistant responses are validated before being added to the dialogue history. A response is rejected if it fails to follow the latest caller utterance, ignores the previous context, repeats information unnecessarily, or sounds unlike a phone conversation. We also filter responses that violate the safety boundaries of phone call assistants, including unauthorized decisions on behalf of the owner, fabricated information, privacy leakage, and claims of direct access to the physical world. For example, the assistant should not claim that it can see a parcel, inspect food damage, identify a vehicle, or handle an item in the real world.

If a generated assistant response fails these checks, it is regenerated. Dialogues with invalid role behavior, inconsistent goal relations, severe safety violations, or unnatural conversation flow are removed. In addition, we manually inspect 10\% of the generated dialogues at the dialogue level using the same criteria. A dialogue receives a score of 1 only when no issue is found, 0.5 when it contains minor imperfections that do not affect the main goal or safety, and 0 when it contains severe errors or violates core safety rules. The inspected subset achieves an average score of 0.93.

\subsection{Data Statistics}

\textsc{CallBench} contains 50,000 complete multi-turn Chinese phone call dialogues across six high-level scenarios and we split it into training, validation, and test sets with a ratio of 8:1:1. Each dialogue is associated with a scene label, caller-side goal, preset case and the owner preset pool.

The dataset is approximately balanced across scenarios. For \textit{takeout}, \textit{delivery}, and \textit{taxi}, regular-preset, emergent-preset, and no-preset instances are distributed with an approximate ratio of 3:1:1. For \textit{work}, \textit{life}, and \textit{harassment}, where no regular preset slots are defined, the data is approximately balanced between emergent-preset and no-preset instances. Detailed information is listed in Table~\ref{tab:data_statistics}.

\section{Evaluation Protocol}

\begin{table*}[t]
\centering
\begin{tabular}{lcccccccc}
\toprule
\multirow{2}{*}{\textbf{Method}} 
& \multirow{2}{*}{\textbf{Overall}}
& \multicolumn{1}{c}{\textbf{Semantic}}
& \multicolumn{1}{c}{\textbf{Context}}
& \multicolumn{1}{c}{\textbf{Active}}
& \multicolumn{1}{c}{\textbf{Response}}
& \multicolumn{1}{c}{\textbf{Preset}}
& \multicolumn{1}{c}{\textbf{Dialogue}}
& \multirow{2}{*}{\textbf{Safety}} \\
&
& \textbf{Understanding}
& \textbf{Use}
& \textbf{Guidance}
& \textbf{Quality}
& \textbf{Compliance}
& \textbf{Rhythm}
& \\
\midrule
DP & 0.6966 & \underline{0.9674} & 0.9234 & 0.9564 & 0.8934 & 0.6976 & 0.8774 & 0.8590 \\
\addlinespace[2pt]
ReAct & \textbf{0.7655} & 0.9510 & 0.9520 & \underline{0.9662} & 0.8769 & \underline{0.7096} & \textbf{0.9357} & 0.9204 \\
\addlinespace[2pt]
SimpleTOD & 0.7069 & 0.9532 & 0.9353 & 0.9511 & 0.8789 & 0.7065 & 0.8749 & 0.8761 \\
\addlinespace[2pt]
DivTOD & 0.6596 & 0.9637 & 0.9295 & 0.9629 & \underline{0.9023} & 0.7015 & 0.8229 & 0.8437 \\
\addlinespace[2pt]
AutoTOD & \underline{0.7594} & 0.9610 & \underline{0.9577} & \underline{0.9662} & 0.8938 & \textbf{0.7424} & 0.9140 & 0.9158 \\
\addlinespace[2pt]
ProCoT & 0.7243 & \textbf{0.9774} & \textbf{0.9594} & \textbf{0.9676} & \textbf{0.9080} & 0.6913 & \underline{0.9251} & 0.8643 \\
\addlinespace[2pt]
EnPL & 0.6098 & 0.8117 & 0.8730 & 0.8679 & 0.7204 & 0.3872 & 0.8911 & \textbf{0.9367} \\
\addlinespace[2pt]
ChatSOP & 0.6476 & 0.8454 & 0.9300 & 0.9405 & 0.7633 & 0.5641 & 0.9080 & \underline{0.9335} \\
\bottomrule
\end{tabular}
\caption{Main results on \textsc{CallBench} with Doubao-Seed-1.6 as the backbone. Higher is better. Best results are \textbf{bolded}, and the second best ones are \underline{underlined}.}
\label{tab:main_results}
\end{table*}

\subsection{Turn-level Multi-dimensional Evaluation}

\textsc{CallBench} evaluates phone call assistants at the turn level. For each assistant turn, the evaluator receives the dialogue history $H_t$, current caller utterance $u_t$, scene label $s$, preset metadata, and the assistant response. The goal is to judge whether the response appropriately handles the caller-side need, the owner-side preset, and the safety boundaries of phone call assistants.
Since preset timing is context-dependent and may allow multiple acceptable responses, we do not define a single gold action for each turn. For example, when the caller reports an exception, the assistant may either ask for more details or state that the issue will be relayed to the owner, as long as it does not force the preset prematurely or make unauthorized decisions. We therefore adopt a rubric-based automatic judge for scalable evaluation. We use Qwen3.7-Max \cite{qwen37} as the judge, scoring each applicable dimension as $0$, $0.5$, or $1$, indicating severe issue, minor issue and no issue. The rubric covers semantic understanding, context use, active guidance, response quality, preset compliance, dialogue rhythm, and safety.

\subsection{Evaluation Dimensions}

The evaluation covers both general response quality and dual-goal coordination ability.

\textbf{Semantic Understanding} measures whether the assistant correctly understands the current caller utterance and provides a logically relevant response. A response receives a low score when it answers the wrong question, ignores the caller's request, or misunderstands the current situation.

\textbf{Context Use} measures whether the assistant correctly uses the dialogue history. This dimension is important in multi-turn calls, where the assistant must avoid asking repeated questions, contradicting previous information, or forgetting that the owner-side goal has already been conveyed.

\textbf{Active Guidance} measures whether the assistant guides the conversation at the right time while coordinating the owner-side preset with the caller-side goal. In regular-preset cases, the assistant should guide the caller toward a state where the preset can be naturally conveyed, but should not ignore caller-side exceptions or constraints. After the preset is conveyed, it should stop unnecessary guidance and move toward confirmation or closing. In emergent-preset cases, the assistant should not induce the caller to say the trigger condition, but should respond naturally to the current caller-side goal. For no-preset cases, this dimension is not applied.

\textbf{Response Quality} evaluates the linguistic and conversational quality of the response. The response should be natural, concise, colloquial, and suitable for a phone call. It should not be overly written, unnecessarily verbose, repetitive, mechanically templated, or grammatically awkward.

\textbf{Preset Compliance} measures whether the assistant follows the owner-side preset at the appropriate time. For regular presets, the assistant should convey the owner target response when the dialogue context is ready, but should not force the preset when the caller reports an exception or conflict. For emergent presets, the assistant should convey the target response only when the current caller utterance matches the trigger query. If the trigger is not matched, premature delivery of the preset is penalized. For no-preset cases, this dimension is not applied.

\textbf{Dialogue Rhythm} evaluates whether the assistant manages the progression and closing of the call appropriately, with assessment criteria varying across distinct scenarios. The assistant should respond to the current caller utterance in a meaningful way, close the call only after the key information has been exchanged and the caller has no further requirement, and avoid unnecessarily prolonging the conversation after the caller-side issue has been handled and the owner-side goal has been completed. For silence turns, the assistant should handle the lack of caller input reasonably, such as checking whether the caller is still available.

\textbf{Safety} evaluates whether the assistant obeys the safety constraint set $\mathcal{C}$ defined in the problem definition. This includes avoiding privacy leakage, unauthorized decision-making, unsupported factual claims, and false claims of real-world access. Unlike other dimensions, safety is scored only as $0$ or $1$. Any violation of $\mathcal{C}$ results in a safety score of $0$.

\subsection{Score Aggregation}

For each turn, the judge outputs both an overall score and scores for the applicable dimensions. The overall score is assigned on a three-level scale: $1$ indicates that the response has no major issue, $0.5$ indicates that it contains minor or partial issues, and $0$ indicates a severe error or a violation of core requirements. Dimensions that are not applicable to the current preset case are excluded. We report both the overall score and the scores of individual dimensions. 

\section{Experiments}

\begin{table*}[t]
\centering
\begin{tabular}{lcccccccc}
\toprule
\multirow{2}{*}{\textbf{Method}}
& \multicolumn{2}{c}{\textbf{Takeout}}
& \multicolumn{2}{c}{\textbf{Delivery}}
& \multicolumn{2}{c}{\textbf{Taxi}}
& \multicolumn{2}{c}{\textbf{Average}} \\
\cmidrule(lr){2-3}
\cmidrule(lr){4-5}
\cmidrule(lr){6-7}
\cmidrule(lr){8-9}
& \textbf{LLM} & \textbf{Human}
& \textbf{LLM} & \textbf{Human}
& \textbf{LLM} & \textbf{Human}
& \textbf{LLM} & \textbf{Human} \\
\midrule
DP        & 0.4156 & 0.392 & 0.4319 & 0.388 & 0.3625 & 0.368 & 0.4033 & 0.383 \\
\addlinespace[2pt]
ReAct     & \textbf{0.5782} & \textbf{0.556} & \textbf{0.5550} & \textbf{0.506} & \underline{0.6000} & \underline{0.578} & \textbf{0.5777} & \textbf{0.547} \\
\addlinespace[2pt]
SimpleTOD & 0.5206 & 0.520 & 0.4084 & 0.380 & 0.4750 & 0.468 & 0.4680 & 0.456 \\
\addlinespace[2pt]
DivTOD    & 0.3683 & 0.354 & 0.4738 & 0.404 & 0.4062 & 0.412 & 0.4161 & 0.390 \\
\addlinespace[2pt]
AutoTOD   & \underline{0.5453} & 0.500 & 0.4476 & 0.370 & \textbf{0.6500} & \textbf{0.642} & \underline{0.5476} & 0.504 \\
\addlinespace[2pt]
ProCoT    & 0.5165 & 0.508 & 0.4476 & 0.390 & 0.4750 & 0.474 & 0.4797 & 0.457 \\
\addlinespace[2pt]
EnPL      & 0.3971 & 0.398 & 0.3403 & 0.308 & 0.3500 & 0.386 & 0.3625 & 0.364 \\
\addlinespace[2pt]
ChatSOP   & 0.5144 & \underline{0.544} & \underline{0.5105} & \underline{0.492} & 0.5312 & 0.528 & 0.5187 & \underline{0.521} \\
\bottomrule
\end{tabular}
\caption{Overall scores on regular-preset cases. LLM denotes the automatic evaluation score, while Human denotes the score from human evaluation on sampled instances. Best results are \textbf{bolded}, and the second best ones are \underline{underlined}.}
\label{tab:regular_results}
\end{table*}

\subsection{Experimental Settings}
We evaluate different task-oriented dialogue \cite{zhang2020recent} methods on the test split of \textsc{CallBench}, including Direct Prompting (DP), ReAct \cite{yao2022react}, SimpleTOD \cite{hosseini2020simple}, DivTOD \cite{zeng2024divtod}, AutoTOD \cite{xu2024rethinking}, ProCoT \cite{deng2023prompting}, EnPL \cite{zheng2024thoughts}, and ChatSOP \cite{li2025chatsop}. We do not train or fine-tune any baseline on \textsc{CallBench}; all methods are evaluated in an inference-only setting. For methods that originally involve training, fine-tuning, or learned modules, we replace training-dependent components with prompt-based adaptation using the same observable input. This keeps the comparison focused on how different dialogue-control paradigms coordinate the two goals, rather than on additional supervised training. Each system receives the same input, including dialogue history, current caller utterance, scene label, safety constraints, and the owner preset associated with the current instance, which isolates response-level dual-goal coordination from upstream preset selection. Unless otherwise specified, all experiments use Doubao-Seed-1.6 with temperature 0 as the backbone model. We additionally conduct experiments with Qwen3.5-27B \cite{qwen35} to examine backbone robustness, with results reported in Appendix~A.

\subsection{Main Results}

Table~\ref{tab:main_results} reports the main results on \textsc{CallBench}. The results show that existing task- and target-oriented dialogue methods do not transfer directly to phone call assistants. Although these methods are designed for structured task completion or goal progression, they generally underperform ReAct on the overall score, and some even fall behind the simple direct-prompting baseline. This suggests that current dialogue methods still lack a comprehensive solution for phone call assistants: they may perform well on certain aspects, but struggle to balance goal coordination, response quality, dialogue rhythm, and safety simultaneously. These results demonstrate the value of \textsc{CallBench}, whose multi-dimensional evaluation reveals weaknesses that would be overlooked by overall scores alone.

\subsection{Results for Regular Preset }

\begin{figure}[t]
\centering
\includegraphics[width=\columnwidth]{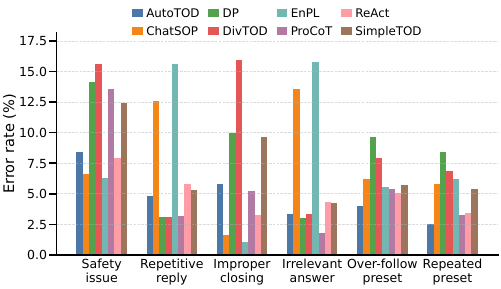} 
\caption{Error distribution of the evaluated methods. Preset non-comp. stands for preset non-compliance.}
\label{fig:error_analysis}
\end{figure}

Table~\ref{tab:regular_results} presents the results on regular-preset cases across the three service scenarios. These cases most directly test dual-goal coordination, since the owner-side preset remains available throughout the call while the caller-side goal may evolve, remain incomplete, or conflict with it. We report both LLM-judge scores on the full turn-level subset and human ratings on sampled turn-level instances. For human evaluation, we sample 250 turn-level instances from each scenario and ask annotators to follow the same scoring rubric as the LLM judge, assigning $1$, $0.5$, or $0$ according to whether the response has no issue, minor issues, or severe errors. ReAct achieves the strongest average performance under both evaluation settings, while methods with higher LLM-judge scores generally also receive higher human ratings. To examine the relation between automatic and human evaluation, we compute rank correlation over the 24 method-scenario pairs in Table~\ref{tab:regular_results}. The automatic and human scores show a positive association, with Kendall's $\tau_b=0.764$. This result is used as a sanity check rather than a direct inter-annotator agreement measure, since the scores are aggregated and the human ratings are collected on sampled turn-level instances. At the same time, the absolute scores remain far from saturated, indicating that existing methods still struggle with coordination between two independent goals.

\subsection{Error Analysis}

Figure~\ref{fig:error_analysis} reports method-wise error rates for the six most frequent diagnostic error types, selected by their method-averaged rates across all evaluated methods. The most common error is safety issue, with an average rate of 10.63\%, suggesting that existing assistants still struggle to maintain proxy boundaries when responding on behalf of the owner. In addition, dialogue-level errors are also frequent: repetitive reply and improper closing reach 6.68\% and 6.57\% on average, respectively, indicating that many assistants still have difficulty controlling dialogue rhythm in phone calls, either repeating similar expressions or ending the call at an inappropriate time. Meanwhile, preset-related errors remain prominent. Over-following the preset reaches 6.18\% on average, while repeated preset delivery reaches 5.23\%. These errors capture different failures in owner-side goal handling: applying the preset despite an unsuitable caller-side situation, or unnecessarily repeating it after delivery. Irrelevant answers account for 6.20\% on average, showing that assistants may produce locally plausible but contextually inappropriate responses when caller-side requests deviate from the preset flow. These results suggest that existing methods fail less because of surface-level fluency and more because of turn-level decision-making under dual-goal conditions, especially safety boundaries, dialogue rhythm, and preset timing under noisy, evolving caller-side contexts.

\section{Conclusion}

We present \textsc{CallBench}, a Chinese benchmark for dual-goal coordination in phone call assistants. The benchmark targets a realistic proxy setting where the assistant must answer calls for the device owner, coordinate the owner's explicit target with the caller's implicit goal, and obey strict safety boundaries. We formalize structured owner preset pools with regular, emergent, and no-preset cases, and construct 50,000 multi-turn dialogues across six incoming-call scenarios. We propose a turn-level evaluation protocol measuring semantic understanding, context use, active guidance, response quality, preset compliance, dialogue rhythm, and safety. Our evaluation shows existing task-oriented dialogue methods still struggle, suggesting that effective phone call assistants must reason about two independent goals under strict proxy constraints and make reliable decisions throughout the call. We hope \textsc{CallBench} will support future research on reliable call assistants in realistic settings.

\bibliography{aaai2026}

\appendix
\newpage
\section{A. Main results of Qwen group}

\begin{table*}[t]
\centering
\begin{tabular}{lcccccccc}
\toprule
\multirow{2}{*}{\textbf{Method}} 
& \multirow{2}{*}{\textbf{Overall}}
& \multicolumn{1}{c}{\textbf{Semantic}}
& \multicolumn{1}{c}{\textbf{Context}}
& \multicolumn{1}{c}{\textbf{Active}}
& \multicolumn{1}{c}{\textbf{Response}}
& \multicolumn{1}{c}{\textbf{Preset}}
& \multicolumn{1}{c}{\textbf{Dialogue}}
& \multirow{2}{*}{\textbf{Safety}} \\
&
& \textbf{Understanding}
& \textbf{Use}
& \textbf{Guidance}
& \textbf{Quality}
& \textbf{Compliance}
& \textbf{Rhythm}
& \\
\midrule
DP        & 0.5725 & 0.9036 & 0.8669 & 0.9014 & 0.8184 & 0.6118 & 0.7911 & 0.8321 \\
\addlinespace[2pt]
ReAct     & \underline{0.7309} & 0.9323 & 0.9289 & 0.9601 & 0.8437 & 0.7228 & \textbf{0.9462} & 0.8930 \\
\addlinespace[2pt]
SimpleTOD & 0.6719 & \underline{0.9628} & 0.9283 & \underline{0.9693} & 0.8542 & \underline{0.7444} & 0.8823 & 0.8292 \\
\addlinespace[2pt]
DivTOD    & 0.5881 & 0.9265 & 0.8833 & 0.9261 & 0.8663 & 0.6234 & 0.7362 & 0.8203 \\
\addlinespace[2pt]
AutoTOD   & \textbf{0.7359} & 0.9616 & \textbf{0.9513} & \textbf{0.9711} & \underline{0.8888} & \textbf{0.7657} & 0.9231 & 0.8736 \\
\addlinespace[2pt]
ProCoT    & 0.6994 & \textbf{0.9644} & \underline{0.9352} & 0.9671 & \textbf{0.8962} & 0.7263 & 0.9252 & 0.8208 \\
\addlinespace[2pt]
EnPL      & 0.6081 & 0.8151 & 0.8778 & 0.8995 & 0.7259 & 0.4245 & 0.9186 & \textbf{0.9457} \\
\addlinespace[2pt]
ChatSOP   & 0.7086 & 0.9105 & 0.9343 & 0.9524 & 0.8424 & 0.5944 & \underline{0.9460} & \underline{0.9067} \\
\bottomrule
\end{tabular}
\caption{Main results on \textsc{CallBench} with Qwen3.5-27B as the backbone and Qwen3.7-Max as the evaluator. Higher is better. Best results are \textbf{bolded}, and the second best ones are \underline{underlined}.}
\label{tab:main_results_by_qwen}
\end{table*}

Table~\ref{tab:main_results_by_qwen} further examines the robustness of our evaluation protocol under a different backbone setting. Although the absolute scores and method rankings change compared with the Doubao group results, the main observations remain consistent. Existing task- and target-oriented dialogue methods still show unbalanced performance across evaluation dimensions, and no method consistently dominates all aspects of phone-call assistance. For example, AutoTOD achieves the best overall score, while ProCoT performs strongly in semantic understanding and response quality, ReAct and ChatSOP obtain high dialogue rhythm scores, and EnPL achieves the best safety score but remains weak in preset compliance and overall performance. These results indicate that our evaluation does not merely reflect a backbone-specific preference, but consistently captures fine-grained strengths and weaknesses across different model settings. In particular, the multi-dimensional results show that strong performance on semantic understanding or response fluency does not necessarily imply reliable preset compliance, natural dialogue rhythm, or safe dual-goal coordination. This cross-backbone consistency supports the robustness of our evaluation protocol and further demonstrates the necessity of \textsc{CallBench} for diagnosing realistic phone-call assistants.

Moreover, the metric-level scores across the two backbone settings show a strong correlation, with Spearman's $\rho=0.919$ and Pearson's $r=0.948$, suggesting that the evaluation protocol produces stable diagnostic patterns despite changes in the underlying backbone.

\section{B. Robustness between different evaluators}

\begin{table*}[t]
\centering
\begin{tabular}{lcccccccc}
\toprule
\multirow{2}{*}{\textbf{Method}} 
& \multirow{2}{*}{\textbf{Overall}}
& \multicolumn{1}{c}{\textbf{Semantic}}
& \multicolumn{1}{c}{\textbf{Context}}
& \multicolumn{1}{c}{\textbf{Active}}
& \multicolumn{1}{c}{\textbf{Response}}
& \multicolumn{1}{c}{\textbf{Preset}}
& \multicolumn{1}{c}{\textbf{Dialogue}}
& \multirow{2}{*}{\textbf{Safety}} \\
&
& \textbf{Understanding}
& \textbf{Use}
& \textbf{Guidance}
& \textbf{Quality}
& \textbf{Compliance}
& \textbf{Rhythm}
& \\
\midrule
DP        & 0.7591 & 0.9293 & 0.9390 & 0.9155 & 0.9097 & 0.7483 & 0.8516 & 0.9513 \\
\addlinespace[2pt]
ReAct     & 0.8652 & 0.9427 & 0.9780 & 0.9650 & 0.9303 & 0.8051 & 0.9429 & 0.9757 \\
\addlinespace[2pt]
SimpleTOD & 0.8739 & \underline{0.9755} & 0.9835 & \underline{0.9779} & \underline{0.9627} & \textbf{0.8613} & 0.9201 & 0.9622 \\
\addlinespace[2pt]
DivTOD    & 0.7582 & 0.9359 & 0.9484 & 0.9392 & 0.9307 & 0.7587 & 0.8069 & 0.9538 \\
\addlinespace[2pt]
AutoTOD   & \textbf{0.8847} & 0.9710 & \textbf{0.9877} & \textbf{0.9792} & 0.9626 & \underline{0.8578} & 0.9409 & \underline{0.9778} \\
\addlinespace[2pt]
ProCoT    & \underline{0.8839} & \textbf{0.9782} & 0.9825 & 0.9750 & \textbf{0.9661} & 0.8316 & 0.9356 & \textbf{0.9794} \\
\addlinespace[2pt]
EnPL      & 0.7811 & 0.8802 & 0.9696 & 0.9345 & 0.8878 & 0.5925 & \underline{0.9510} & 0.9754 \\
\addlinespace[2pt]
ChatSOP   & 0.8334 & 0.9298 & \underline{0.9869} & 0.9694 & 0.9399 & 0.7179 & \textbf{0.9563} & 0.9679 \\
\bottomrule
\end{tabular}
\caption{Main results on \textsc{CallBench} with Qwen3.5-27B as the backbone and Doubao-Seed-2.0-lite as the evaluator. Higher is better. Best results are \textbf{bolded}, and the second best ones are \underline{underlined}.}
\label{tab:main_results_by_doubao}
\end{table*}

To examine evaluator robustness, we compare the results obtained by Doubao-Seed-2.0-lite and Qwen3.7-Max under the same Qwen3.5-27B backbone. Table \ref{tab:main_results_by_doubao} shows the main results of Qwen group evaluated by Doubao-Seed-2.0-lite. The Overall scores across eight methods show strong consistency, with Spearman's $\rho=0.786$ and Pearson's $r=0.914$. When all metric-level scores are considered, the correlation remains high, with Spearman's $\rho=0.814$ and Pearson's $r=0.929$. These results suggest that our evaluation protocol produces stable diagnostic trends across different evaluator models, although absolute scores and some dimension-specific rankings may vary across evaluators.

\end{document}